\documentclass{article} % For LaTeX2e
\usepackage{nips14submit_e,times}
\usepackage{hyperref}
\usepackage{url}
\usepackage{amsfonts}
\usepackage{amsmath}
\usepackage{graphicx}

\newcommand{\comment}[1]{}
\title{EquiNMF: Graph Regularized Multiview Nonnegative Matrix Factorization}

\author{
Daniel Hidru\thanks{DH and AG were supported by the SickKids Foundation} \\
SickKids Research Institute\\
686 Bay St, Toronto, ON, Canada\\
\texttt{danielhidru@hotmail.com} \\
\And
Anna Goldenberg\\
SickKids Research Institute, University of Toronto\\
686 Bay St, Toronto, ON, Canada\\
\texttt{anna.goldenberg@utoronto.ca}\\
}

\nipsfinalcopy % Uncomment for camera-ready version

\begin{document}

\maketitle

\begin{abstract}
Nonnegative matrix factorization (NMF) methods have proved to be powerful across a wide range of real-world clustering applications. Integrating multiple types of measurements for the same objects/subjects allows us to gain a deeper understanding of the data and refine the clustering. We have developed a novel Graph-reguarized multiview NMF-based method for data integration called EquiNMF. The parameters for our method are set in a completely automated data-specific unsupervised fashion, a highly desirable property in real-world applications. We performed extensive and comprehensive experiments on multiview imaging data. We show that EquiNMF consistently outperforms other single-view NMF methods used on concatenated data and multi-view NMF methods with different types of regularizations. 
%Furthermore, our EquiNMF works with partially missing data without the need for preliminary data %imputation making our approach appealing to a broader range of applications.
\end{abstract}

\section{Introduction}
Combining multiple sources of evidence helps us gain a deeper understanding of the data. If unsupervised clustering is to be performed, a simple way to utilize the multiple sources of data is to concatenate them after normalizing their features and to perform clustering on the unified data set. This is not an ideal strategy because concatenation is likely to cause the loss of structure inherent in individual datasets which could compromise the identification of clusters. For this reason, methods have been developed to cluster data sets preserving their multiview structure (e.g. ~\cite{chen2010predictive}).

Nonnegative Matrix Factorization (NMF) has achieved wide spread popularity and has become a clustering method of choice in many applications, such as imaging \cite{guillamet2002analyzing}, blind-source separation \cite{virtanen2007monaural} and computational biology \cite{wang2013non}. With NMF, clustering is performed on the lower dimensional representation of the data which arises from the matrix factors. The power of the method lies in the quality of the latent embedding which was shown to yield superior performance to PCA~\cite{lee1999learning}. Many NMF variants have been proposed to improve the performance~\cite{Cichocki09}. For example, sparsity constraints have been enforced to identify better bases for NMF~\cite{hoyer2004non}. Graph regularization has also been added to generate superior clustering results~\cite{cai2011graph}.

% Similarity graph contains a global information of manifold structures. With proper graph regularization, many learning systems can take consideration of underlying cluster structures~\cite{zheng2011graph}. 

Many application areas are now interested in data integration since integrating various sources of data can yield a much finer picture of the domain. A recently proposed MultiNMF~\cite{liu2013multi} extends NMF to the multi-view clustering problem, by constraining each view's lower dimensional representation to be similar to each other. The current MultiNMF has a major disadvantage that it does not capture the geometric structure of the data, which has been shown to improve NMF for single views ~\cite{cai2011graph}. We propose EquiNMF: a graph-based regularized multi-view method where the parameters are automatically learned from data. It results in significant performance improvements over four alternative approaches on three imaging datasets and shows consistency and robustness across a variety of parameter settings that in our case determine relative contributions of multiple views. Importantly, while competing methods perform well on one dataset and badly on others, our approach is able to deal with the dataå diversity appropriately.

%As a method of broad applicability, the majority of the NMF and consequently MultiNMF approaches and their implementations are missing a key property - ability to deal with missing data. Most real world applications contain missing data: in imaging, it is a partial obstruction of a view, in computational biology and other empirical data sciences, the physical measurements are missing due to technical difficulties and stringent quality control filters. It has been previously shown and our experiments confirm that data imputation performed prior to applying a given method results in suboptimal performance \cite{donders2006review}. In our work we address this problem as part of our EquiNMF approach. Here, we illustrate two mechanisms for dealing with missing data problem and discuss their pitfalls and benefits. 

Our three major contributions are 1) a novel formalization of a graph regularized multi-view NMF which results in much improved accuracy; 2) reformulating the multi-view objective to simplify and reduce the complexity of the approach by explicitly representing equal view contribution without the consensus matrix; 3) automatic parameter estimation in a truly unsupervised setting.

\section{Overview of Nonnegative Matrix Factorization and relevant extensions}
\subsection{Nonnegative Matrix Factorization}
Nonnegative Matrix Factorization (NMF) is a method used to factorize a matrix of nonnegative entries into the product of two lower dimensional, nonnegative matrices. Let $X \in \mathbb{R}_{+}^{M \times N}$, where $X$ contains $N$ data points and $M$ nonnegative measurements for each data point. NMF attempts to find $U \in \mathbb{R}_{+}^{M \times K}$ and $V \in \mathbb{R}_{+}^{N \times K}$ such that $X \approx UV^T$~\cite{lee1999learning}.  This task is expressed mathematically as the following optimization problem with iterative updates~\cite{seung2001algorithms}: 
\begin{equation}\label{NMF}
\min_{U,V \geq 0} {|| X - U{V}^T||_{F}^2}; \quad\quad
U_{i,k} \leftarrow U_{i,k}\frac{(XV)_{i,k}}{(U{V}^TV)_{i,k}},\quad
V_{j,k} \leftarrow V_{j,k}\frac{({X}^TU)_{j,k}}{(V{U}^TU)_{j,k}}
\end{equation}
\subsection{Graph Regularized NMF}
Graph Regularized NMF (GNMF) is an extension of NMF which has been shown to improve the quality of the factorization of $X$ ~\cite{cai2011graph}. This improvement has been achieved through the addition of a regularization term which causes similar data points to have similar lower dimensional representation. This in turn reduces overfitting of the basis vectors. 

Let $W$ be an $N \times N$ symmetric matrix representing the similarity between the $N$ data points. Let $D$ be the diagonal matrix such that $D_{jj}=\sum_l W_{jl}$, then the Laplacian of $W$ is $\Delta = D-W$. GNMF attempts to solve the following optimization problem with iterative updates~\cite{cai2011graph}:
\begin{eqnarray}\label{GNMF Objective}
\min_{U,V \geq 0}  {|| X - U{V}^T||_{F}^2 + \gamma Tr(V^T\Delta V)};\nonumber\\
U_{i,k} \leftarrow U_{i,k}\frac{(XV)_{i,k}}{(U{V}^TV)_{i,k}}, 
V_{j,k} \leftarrow V_{j,k}\frac{({X}^TU)_{j,k} + \gamma {(WV)}_{j,k}}{(V{U}^TU)_{j,k} + \gamma {(DV)}_{j,k}}
\end{eqnarray}
\subsection{Multi-view NMF}
Multi-view NMF (MultiNMF) is an extension of NMF to multiple nonnegative matrices describing the same set of data points. Let $\{X^{(1)},...,X^{(n_v)}\}$ be $n_v$ views of a set of data points. MultiNMF attempts to approximate  $X^{(v)} \approx U^{(v)}(V^{(v)})^T$ for each $v$, while the constraining the $V^{(v)}$'s to be similar ~\cite{liu2013multi}. This is achieved by solving the following optimization problem:

\begin{equation}\label{MultiNMF Objective}
\min_{U^{(v)},V^{(v)},V^* \geq 0} \sum_{v=1}^{n_{v}}{|| X^{(v)} - U^{(v)}({V}^{(v)})^T||_{F}^2} + \sum_{v=1}^{n_{v}}\lambda_{v}{|| V^{(v)}Q^{(v)} - V^*||_{F}^2}
\end{equation}

In the optimization above, $Q^{(v)}$ is a matrix which constrains the column sums of $U^{(v)}$ to make the $V^{(v)}$'s comparable~\cite{liu2013multi}. The multi-view data is reduced to $V^*$.

\section{EquiNMF: Equivalent-contribution multiView NMF}
Capturing internal structure of the data within each view in a multiview problem is key to improving performance and gaining meaningful insight into the data and its underlying domain (e.g. \cite{Wang:2014}).  We thus propose a novel graph-regularized multi-view approach. The usual problem of the multi-view setting, especially in the unsupervised scenario, is that it is not clear how to chose how much each view should contribute to the final objective. The selection of parameter values in the objective function has a substantial effect on the results of NMF methods which require them. Previous methods determined these values empirically using their labeled data and recommended the use of the same parameter values on all datadsets. Since the appropriate parameter values may depend on the size and scale of the data being used, we have developed a method to determine these parameters  from the data by assuming equivalent contributions of each view (note that it does not mean that each view gets the same coefficient as is done in many multi-view approaches). 

\subsection{Graph Regularized Multi-view NMF}
Here we show how to extend graph-regularized NMF (GNMF) to the multi-view setting. Let $\{X^{(1)},...,X^{(n_v)}\}$ be $n_v$ views of a set of $N$ data points, such that $X^{(v)} \in \mathbb{R}_{+}^{M_{v} \times N}$.  The proposed method attempts to approximate $X^{(v)} \approx U^{(v)}V^T$ for each $v$, where $U^{(v)} \in \mathbb{R}_{+}^{M_v \times K}$ and $V \in \mathbb{R}_{+}^{N \times K}$ and the coefficient matrix $V$ is shared between all of the views.

Since $V$ is shared between all of the views, we would like to guarantee that the entries from each row of $V$ have a magnitude which will allow them to approximate the corresponding column in each of the views. Suppose that $X \approx UV^T$, $|| X_{.,j} ||_1 = 1$ and $|| U_{.,k} ||_1 = 1$ for each $k$. Then:

\begin{equation} \label{Normalization Justification}
1 = || X_{.,j} ||_1 \approx \sum_{k=1}^K || U_{.,k} V_{j,k} ||_1 = \sum_{k=1}^K || V_{j,k} ||_1 = || V_{j,.} ||_1
\end{equation}

Given the above constraints, a single $V$ can be used to approximate each of the views simultaneously.  This motivates us to normalize the original data such that $|| X_{.,j}^{(v)} ||_1 = 1$ and express the other constraints within the optimization problem below:
\begin{equation}\label{EquiNMF Objective}
\min_{U^{(v)},V \geq 0} \sum_{v=1}^{n_{v}}\alpha_{v}|| X^{(v)} - U^{(v)}C^{(v)}V^T||_{F}^2 +\gamma{Tr(V^T\Delta V)}
\end{equation}
Where  $C^{(v)} = Diag(\frac{1}{\sum_{i=1}^{M_v}U_{i,1}^{(v)}},...,\frac{1}{\sum_{i=1}^{M_v}U_{i,K}^{(v)}})$
is used to constrain the column sums of $U^{(v)}$, as
$||(UC)_{.,k} ||_1 = \sum_{i=1}^{M} (UC)_{i,k} = C_{k,k} \sum_{i=1}^{M} U_{i,k} = 1$.

To solve the optimization problem in (Eq.\ref{EquiNMF Objective}), we derive alternating updates in the same manner as previous NMF papers~\cite{lee1999learning}. First, we fix $V$ and minimize the objective for each $U^{(v)}$. When $V$ is fixed, each of the $U^{(v)}$'s do not depend on each other. For this reason, the $v$ indices have been removed for notational convenience.

For each $U$, we only need to minimize the terms in the objective which depend on it. Let $\Psi$ be the Lagrange multiplier matrix for the constraint $U \geq 0$. Considering the terms which are only relevant to $U$, minimizing the objective is equivalent to minimizing the Lagrange:
\begin{eqnarray} \label{Lagrange of U}
L_U &=& \alpha Tr(UCV^TVC^TU^T - 2XVC^TU^T) + Tr(\Psi U) \nonumber \\
&=& \alpha \sum_{i=1}^{M} ((UCV^TVC^TU^T)_{ii} - 2(XVC^TU^T)_{ii}) + Tr(\Psi U) \nonumber \\
&=& \alpha \sum_{i=1}^{M} \sum_{k=1}^{K} ((UCV^TV)_{ik} - 2(XV)_{ik}) \frac{U_{ik}}{\sum_{l=1}^{M} U_{lk}} + Tr(\Psi U)
\end{eqnarray}
Taking the partial derivative of $L_U$ with respect to $U_{ik}$ gives:
\begin{equation}\label{Partial Derivative of L wrt U}
\frac{\partial L_U}{\partial U_{i,k}} = 2C_{kk} \alpha ((UCV^TV)_{i,k} - \sum_{l=1}^{M} (UCV^TV)_{l,k} (UC)_{l,k} - (XV)_{i,k} + \sum_{l=1}^{M} (XV)_{l,k} (UC)_{l,k}) + \Psi
\end{equation}
If we assume that $U$ was column normalized before the update, then $C = I$.  Using the KKT conditions $\Psi_{i,k} U_{i,k} = 0$ and $\frac{\partial L_U}{\partial U_{i,k}} = 0$, we get the update:

\begin{equation} \label{EquiNMF U Update}
U_{i,k} \leftarrow U_{i,k} \frac{(XV)_{i,k} + \sum_{l=1}^{M} (UV^TV)_{l,k} U_{l,k}}{(UV^TV)_{i,k} + \sum_{l=1}^{M} (XV)_{l,k} U_{l,k}}
\end{equation}

To compute the update for $V$, we first normalize the columns of U.  This normalization does not change the value of the objective and reduces $C$ to the identity matrix. Let $\Phi$ be the Lagrange multiplier matrix for the constraint $V \geq 0$. If we fix each $U^{(v)}$ and only consider the terms which are relevant to $V$, minimizing the objective is equivalent to minimizing the Lagrange:
\begin{equation} \label{Langrange of V}
L_V = \sum_{v=1}^{n_v} \alpha_v (Tr(V(U^{(v)})^TU^{(v)}V^T) - 2Tr((X^{(v)})^TU^{(v)}V^T)) + \gamma Tr(V^T\Delta V) + Tr(\Phi V)
\end{equation}
Taking the derivative of $L_V$ with respect to $V$ gives:
\begin{equation}\label{Derivative of L wrt V}
\frac{\partial L_V}{\partial V} = \sum_{v=1}^{n_v} 2 \alpha_v ((V(U^{(v)})^TU^{(v)}) - ((X^{(v)})^TU^{(v)})) + 2 \gamma \Delta V + \Phi
\end{equation}
Using the KKT conditions $\Phi_{j,k} V_{j,k} = 0$ and $\frac{\partial L_V}{\partial V_{j,k}} = 0$, we get the update:
\begin{equation} \label{EquiNMF V Update}
V_{jk} \leftarrow V_{jk} \frac{\sum_{v=1}^{n_v} \alpha_v((X^{(v)})^TU^{(v)})_{jk} + \gamma (WV)_{jk}} {\sum_{v=1}^{n_v} \alpha_v(V(U^{(v)})^TU^{(v)})_{jk} + \gamma (DV)_{jk}}
\end{equation}

\subsection{Parameter settings}

In an unsupervised multi-view setting, it is reasonable to desire each view to contribute equally to the final result ($V$) unless prior information is available. Each view can be said to contribute equally to the final result if it contributes equally to each intermediate result ($V$ after every update). Since each view contributes to the interemediate result according to the magnitude of the term associated with it in the numerator of Eq.\ref{EquiNMF V Update}, equal contribution of the views can be enforced by having the average contribution of each view to be the same. Since
\begin{eqnarray}\label{Average Contribution of Each View}
E[\alpha_{v}(X^TU)_{j,k}] = \alpha_{v}\sum_{i=1}^{M}E[X_{i,j}U_{i,k}]
\approx \alpha_{v}ME[X_{i,j}]E[U_{i,k}]
= \alpha_{v}M(1/M)(1/M)  = \alpha_{v}/M
\end{eqnarray}
then setting $\alpha_v = M_v$ will ensure that each view contributes equally to the final result.

The selection of the regularization parameter $\gamma$ is also required. If $\gamma$ is too large, then the graph regularization term dominates that might not lead to a desirable effect: data points would be forced to have similar values in $V$, even if this provided a poor approximation. If $\gamma$ is too small, then the graph would have little effect on the result. We thus hypothesize that it is reasonable to set the graph to have the same scale of influence as the data. Since the data has an expected total contribution of $n_v$ with the above parameter setting and 
\begin{eqnarray}\label{Average Contribution of Graph}
E[\gamma(WV)_{j,k}] = \gamma \sum_{l=1}^{N}E[W_{j,l}V_{l,k}]
\approx \gamma N E[W_{j,l}]E[V_{l,k}]
\approx  \gamma N E[W_{j,l}]/K
\end{eqnarray}
then setting $\gamma = n_v * K / (N * E[W_{j,l}])$ will ensure that the graph contributes equally to the final result.

\section{Results}

We have applied EquiNMF to three imaging datasets (Digits, Faces and Butterflies) and compared to four competing approaches (K-means, NMF, GNMF and MultiNMF) using accuracy and normalized mutual information (NMI) \cite{liu2013multi}. 

\subsection{Data description}

A brief description of the three image data sets used in the tests is provided below and the summary of the dimensions can be found in Table~\ref{tab:data}:

\begin{itemize}
\item UCI Handwritten Digits\footnote{\url{http://archive.ics.uci.edu/ml/datasets/Multiple+Features}}: This UCI repository dataset contains handwritten digits from 0 to 9. Each class contains 200 examples. The first view contains 76 Fourier coefficients of the character shapes and the second view contains 240 pixel averages in $2 \times 3$ windows.

\item ORL Face data set: This data set from the ORL database contains images of 40 individuals. The database contains 10 different photos for each individual. The images are grayscale and have been normalized to $112 \times 92$ pixels. The first view contains the raw pixel values and the second view contains GIST ~\cite{oliva2001modeling}.

\item Butterfly data set: This data set contains 10 different classes of butterflies ~\cite{Wang:2009}. Each class contains 55 to 100 images with 832  butterflies in total. The views were formed using two different encodings of the images which describe different statistics of the codebooks. The two encoding methods are Fisher Vector (FV)~\cite{Perronnin:2010} and Vector of Linearly Aggregated Descriptors (VLAD)~\cite{Jegou:2010} with dense SIFT~\cite{Bosch:2007}.
\end{itemize}
\vspace{-.7cm}
\begin{table}[h!]
\caption{Summary of Datasets}
\label{tab:data}
\vskip 0.15in
\begin{center}
\begin{small}
\begin{sc}
\begin{tabular}{lcccr}
\hline
Data set & Samples & Clusters & Features \\
\hline
Digit & 2000& 10& (76, 240) \\
Face & 400& 40& (4096, 59) \\
Butterfly & 832& 10& (10240, 6400) \\
\hline
\end{tabular}
\end{sc}
\end{small}
\end{center}
\vskip -0.1in
\end{table}

\subsection{Experimental Settings}
Each method relied on a random initialization, so each test was performed 20 times. The reduced dimension $K$ of the factor matrices was set to the number of clusters in each data set as in \cite{liu2013multi}. All of the methods which relied on regularization parameters had these parameters set to their recommended values. We use a 5 nearest neighbour similarity matrix to obtain a graph for each view as in ~\cite{cai2011graph}. $W$ was set to the sum of each view's similarity graph.

\subsection{Factor Initialization}
Each of the methods tested had their own form of initialization contained within their code. Our method used a similar style of initialization as MultiNMF \cite{liu2013multi}. The factors were generated from the Uniform[0, 1] distribution and scaled so that the column sums of each $U^{(v)}$ and the row sums of $V$ were set to 1. Then, in a consecutive sequence which cycled through the views 50 times, each $U^{(v)}$ was used for a single iteration of NMF.

\subsection{Method comparisons}
To evaluate our method, we compare its perfomance to the following algorithms:
\begin{itemize}
\item Kmeans: The data is normalized so that $|| X_{.,j} ||_2 = 1$ and concatenated into a single view. Kmeans was performed on the concatenation.

\item Concatenated NMF (NMF): The data is normalized so that $|| X_{.,j} ||_2 = 1$ and concatenated into a single view. NMF is performed on the concatenation.

\item Concatenated GNMF (GNMF): The data is normalized so that $|| X_{.,j} ||_2 = 1$ and concatenated into a single view. GNMF is performed with the recommended value of $\gamma = 100$ ~\cite{cai2011graph}.

\item Multi-view NMF (MultiNMF): The data is normalized so that $|| X||_1 = 1$. MultiNMF is performed with the recommended value of $\lambda = 0.01$ ~\cite{liu2013multi}.

\end{itemize}

To cluster our NMF results, k-means clustering was performed on $V^*$ for MultiNMF and on $V$ for all other methods. Clustering was run with 20 repeats and 100 iterations per repeat.

\begin{table}[h!]
\caption{Clustering accuracy on three imaging datasets. Statistically significantly better performers are in bold (ttest $\alpha=0.05$).}
\label{Clustering accuracy}
\vskip 0.15in
\begin{center}
\begin{small}
\begin{sc}
\begin{tabular}{lcccr}
\hline
Algorithm & Digit & Face & Butterfly \\
\hline
Kmeans & 0.90 $\pm$ .04 & 0.51 $\pm$ .02 & 0.68 $\pm$ .04 \\
NMF & 0.84 $\pm$ .03 & 0.30 $\pm$ .02 & 0.57 $\pm$ .03 \\
GNMF & {\bf0.92} $\pm$ .06 & 0.43 $\pm$ .02 & 0.62 $\pm$ .06 \\
MultiNMF & 0.87 $\pm$ .01 & 0.55 $\pm$ .04 & 0.67 $\pm$ .03 \\
EquiNMF & {\bf 0.93} $\pm$ .04 & {\bf 0.57} $\pm$ .02 & {\bf 0.71} $\pm$ .03 \\
\hline
\end{tabular}
\end{sc}
\end{small}
\end{center}
\vskip -0.1in
\end{table}

\begin{table}[h!]
\caption{Clustering nmi on three imaging datasets. Statistically significantly better performers are in bold (ttest $\alpha=0.05$).}
\label{Clustering nmi}
\vskip 0.15in
\begin{center}
\begin{small}
\begin{sc}
\begin{tabular}{lcccr}
\hline
Algorithm & Digit & Face & Butterfly \\
\hline
Kmeans & 0.83 $\pm$ .01 & 0.73 $\pm$ .02 & 0.68 $\pm$ .02 \\
NMF & 0.78 $\pm$ .02 & 0.54 $\pm$ .01 & 0.52 $\pm$ .03 \\
GNMF & {\bf 0.93} $\pm$ .02 & 0.66 $\pm$ .01 & 0.67 $\pm$ .03 \\
MultiNMF & 0.79 $\pm$ .01 & 0.75 $\pm$ .02 & 0.64 $\pm$ .02 \\
EquiNMF & 0.89 $\pm$ .01 & {\bf 0.83} $\pm$ .01 & {\bf 0.70} $\pm$ .01 \\
\hline
\end{tabular}
\end{sc}
\end{small}
\end{center}
\vskip -0.1in
\end{table}

We observe that NMF used on the concatenated views performs consistently the worst of the compared methods across all 3 datasets. We hypothesize that this is due to the fact that it does not account at all for the internal geometric structure of the data. Interestingly, classic Kmeans performs well outperforming NMF and MultiNMF on Digits and Butterflies. It additionally outperforms GNMF on Faces and Butterfly datasets. Kmeans is a reasonable performer because it takes into account distances in the high dimensional space, something that a single view NMF might miss, but falls short of the best performance since it does not take into account the dependency between measurements. GNMF shows unstable performance, performing very well on Digits, but falling far behind other methods on other datasets. This is due to the fact that as a single view method it cannot use multiple representations of the data effectively. EquiNMF performs consistently better than  all of its competitors except for GMNF on the Digits dataset according to the NMI score (it is significantly better than GNMF according to accuracy). 

\subsection{Parameter selection and robustness}
\begin{figure}[!h]
\centering
\includegraphics[height=9cm]{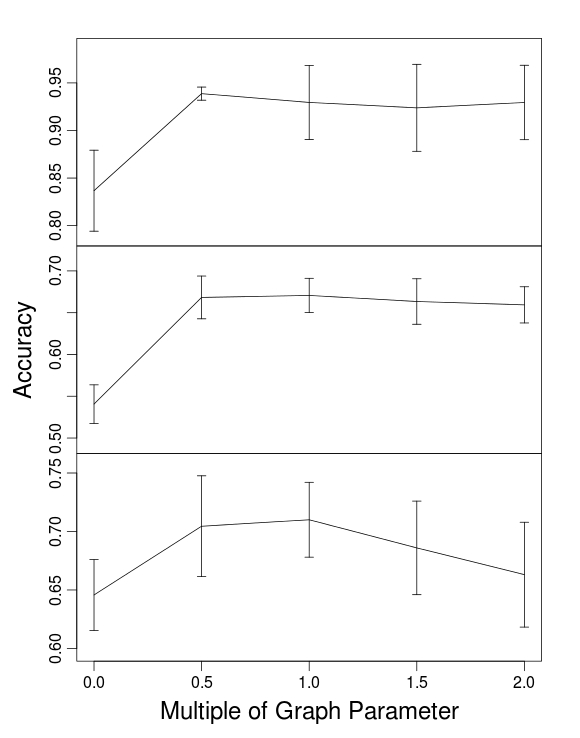}
\caption{EquiNMF clustering accuracy as a function of the coefficient of the selected graph parameter representing its contribution to $V$}
\label{fig:param}
\end{figure}

We plotted the performance of EquiNMF as a function of  a multiplicative constant of the selected graph-regularization parameter $\gamma_v$. Figure \ref{fig:param} shows that EquiNMF is robust for a range of $\gamma_v$ values. The resulting accuracy depends on the contribution of the objective and the regularizer, the graph laplacian in our case. As such, it is very important to set the contribution of the regularization to the right scale. Here, we propose to have comparable contributions of the objective and regularizer, unless prior information is available. Figure \ref{fig:param} shows that while no graph regularization results in significantly worse performance, the equal contribution (multiple of the graph parameter is 1)or half of the objective contribution (mulitple of the graph parameter is 1) perform as well as the best performing parameter setting.  We have also observed that the performance deteriorates once graph regularization is given too much weight (Butterflies, multiplier is equal to 2). We thus recommend to use our automatic setting of equal contribution (multiplier equals 1), resulting in a completely automatically set parameters for EquiNMF in a fully unsupervised though data-specific fashion.

\section{Discussion}

In this paper we propose a graph-regularized multi-view NMF with equal contribution from the views. We have initially extended MultiNMF to use graph regularization. This approach raised a lot of questions, such as should we regularize each view or the consensus matrix or both? Does it matter whether we converge for each U and V before we update the consensus matrix $V^*$? (It turned out that the answer to this question, was yes). Importantly, there was a lot of ambiguity about how to weigh the contributions of each of the views, consensus and each of the potential graph regularizers. We have extensively studied this idea first and found that some of the solutions had substantially increased the performance of MultiNMF, but made search for the best parameter setting very difficult and often impossible without known labels. We have not pursued this approach, since it is not useful in the real world applications where we would ultimately want our method to be used.
\comment{
\begin{eqnarray}\label{MultiGNMF Objective Function}
\min_{U,V,V^* \geq} \sum_{v=1}^{n_{v}}(\alpha_{v}|| X^{(v)} - U^{(v)}({V}^{(v)})^T||_{F}^2
+\gamma_{v}{Tr((V^{(v)}Q^{(v)})^TL^{(v)}V^{(v)}Q^{(v)})}
+ \lambda_{v}{|| V^{(v)}Q^{(v)} - V^*||_{F}^2})
\end{eqnarray}

\begin{equation}\label{MultiGNMF U Update}
U_{i,k} \leftarrow U_{i,k}\frac{\alpha_{v}(XV)_{i,k} + \lambda_{v}\sum_{j=1}^NV_{j,k}V_{j,k}^* + \gamma_{v}(V^TWV)_{kk}}{\alpha_{v}(U{V}^TV)_{i,k} + \lambda_{v}\sum_{j=1}^NV_{j,k}^2 + \gamma_{v}(V^TDV)_{kk}}
\end{equation}

\begin{equation}\label{MultiGNMF Normalization}
U \leftarrow UQ^{-1}, V \leftarrow VQ
\end{equation}

\begin{equation}\label{MultiGNMF V Update}
V_{j,k} \leftarrow V_{j,k}\frac{\alpha_{v}(X^TU)_{j,k} + \lambda_{v}V_{j,k}^* + \gamma_{v}(WV)_{jk}}{\alpha_{v}(V{U}^TU)_{j,k} + \lambda_{v}V_{j,k} + \gamma_{v}(DV)_{jk}}
\end{equation}

\begin{equation}\label{MultiGNMF V* Update}
V^* = \frac{\sum_{v=1}^{n_v}\lambda_vV^{(v)}}{\sum_{v=1}^{n_v}\lambda_v}
\end{equation}
}

Our EquiNMF has many advantages over the graph-regularized MultiNMF approach. For example, automatically setting the parameters of the graph-regularized MultiNMF by using our assumption of equal view contribution is not fully transferable to MultiNMF because there is no way to determine the appropriate proportion of influence that $V^*$ should have on each $V$. Additional advantage of using EquiNMF is that without the consensus, there is no longer a need to determine the order of updates.  In MultiNMF, each $U$, $V$ pair are updated till convergence before $V^*$ is updated. Regularizing $V$ towards a consensus or average is bad.  In theory, as the regularization parameter increases, the method is equivalent to concatenation.  This is bad because concatenation does not allow of the equal contribution of views to the determination of $V$. In practice, as the regularization parameter increases, the $V$'s are similar, but are not necessarily a good approximation of the data. Due to the constraint, it is more difficult to move them from their initialization.

Some other interesting observations about EquiNMF that we found from our extensive experiments are for example, that  constraining (normalizing) the length of rows and columns. Under the constraints on $X$ and $U$ which we imposed above, $|| V_{j,.} ||_1 \approx 1$. In this case, we may wish to impose the row constraint $|| V_{j,.} ||_1 =l 1$ in a similar manner to the column constraints imposed on $U$. Unfortunately, this causes a deterioration in performance, as the model becomes over constrained and loses its expressiveness.  

Initialization also plays an important role. We found that initializing the matrices with (s)kmeans + noise does not allow the method to improve on the initialization. We have observed that our method performs well even with random initialization but has high variance in performance and thus we recommend to use our proposed initialization as it does not add a heavy computational load to the method.

Finally, in an unsupervised multiview setting the $\alpha$ parameters cannot be determined by cross validation, as each view's error would decrease as their parameter, and influence on $V$, increased.  The graph parameter $\gamma$ may be determined by cross validation, but this is not necessary because of our heuristic. If the graph parameter is determined by cross validation, our heuristic gives a reasonable scale to select parameters from.

\section{Conclusion}

Many application areas of machine learning are now looking for multiview methods that will help domain experts to gain deeper understanding of their data. Being a powerful paradigm, NMF has received a wide acclaim in many application areas and thus it is of practical importance to develop novel multiview NMF methods. Existing multiview NMF methods have all relied on supervised parameter detection, either through simulations or through real-world datasets where labels are available. Here we are making two major contributions to the field: 1)  a novel graph-regularized multi-view method that outperforms its state-of-the-art competitors; 2) an automatic way to set all the parameters for our model in unsupervised data-specific fashion. We hope that our approach will be of wide applicability in multiview settings. We will provide both R and matlab code upon acceptance.

\bibliographystyle{plain}
\bibliography{nmf_nips2014}

\end{document}